\documentclass[letterpaper]{article} 
\usepackage{aaai2027}  
\usepackage[hyphens]{url}  

\usepackage{graphicx} 
\usepackage{natbib}  
\usepackage{caption} 
\usepackage{algorithm}
\usepackage{algorithmic}
\usepackage{url}
\usepackage{newfloat}
\usepackage{listings}
\DeclareCaptionStyle{ruled}{labelfont=normalfont,labelsep=colon,strut=off} 
\floatstyle{ruled}
\newfloat{listing}{tb}{lst}{}
\floatname{listing}{Listing}

\usepackage{booktabs}

\usepackage{array}
\nocopyright 

\title{Harnessing VLM in Embodied RL: From Reward Regressor to Success Detector via Exploration-Augmented Key-Observation Gating}
\author {
    Xin Liu\textsuperscript{\rm 1,\rm 2}\equalcontrib,
    Yutong Jiang\textsuperscript{\rm 1, \rm 2}\equalcontrib,
    Jian Su\textsuperscript{\rm 3},
    Ziheng Wang\textsuperscript{\rm 3},
    Chuang Wang\textsuperscript{\rm 3}, \\
    Haoran Li\textsuperscript{\rm 1,\rm 2}\corresponding, 
    Dongbin Zhao\textsuperscript{\rm 1,\rm 2}
}
\affiliations {
    \textsuperscript{\rm 1}State Key Laboratory of Multimodal Artificial Intelligence Systems, Institute of Automation, Chinese Academy of Sciences\\
    \textsuperscript{\rm 2}School of Artificial Intelligence, University of the Chinese Academy of Sciences\\
    \textsuperscript{\rm 3}AGIBOT\\
    liuxin21a@mails.ucas.ac.cn, jiangyutong236@mails.ucas.ac.cn, lihaoran2015@ia.ac.cn
}

\begin{document}

\maketitle

\begin{abstract}
VLMs provide promising solutions to embodied reward design. 
Most current methods that train VLMs as reward regressors not only require extensive annotations but are also primarily designed for finetuning pretrained policies, and perform less well in the more challenging from-scratch RL setting.
This issue stems from a fundamental mismatch: regressing precise rewards is misaligned with what VLMs naturally excel at. We posit that it is more aligned with VLMs’ inherent multi-modal VQA capabilities for the models to function as success detectors,
and propose \textbf{VLM success reward via Exploration-Augmented Key-Observation Gating (VLMeg)} to well harness VLM for embodied reward design. As the core of VLMeg, the proposed exploration-augmented key-observation gating first determines whether the current RL episode is judgment-worthy based on its similarity to a generated success trajectory. Then, the key observation with the lowest confusion from the judgment-worthy episode will be selected for VLM evaluation as episodic reward. This gating mechanism not only avoids redundant calls but also enables higher detection accuracy of VLM. Considering the sparsity of judgment-worthy episodes, the advanced exploration incentives based on the generated trajectory are introduced, enabling more efficient exploration toward judgment-worthy trajectories for more VLM reward triggers and thus gaining higher RL efficiency. Extensive experiments in MetaWorld demonstrate VLMeg achieves state-of-the-art from-scratch RL, outperforming advanced VLM-based baselines on 12 challenging reward-free tasks. Furthermore, VLMeg also enables effective RL without any ground-truth reward in 3 challenging real-world experiments.
Code will be open after acceptance at https://github.com/Raina915/VLMeg.
\end{abstract}

\section{Introduction}
\begin{figure}[!t]
\centering
\includegraphics[width=0.99\linewidth]{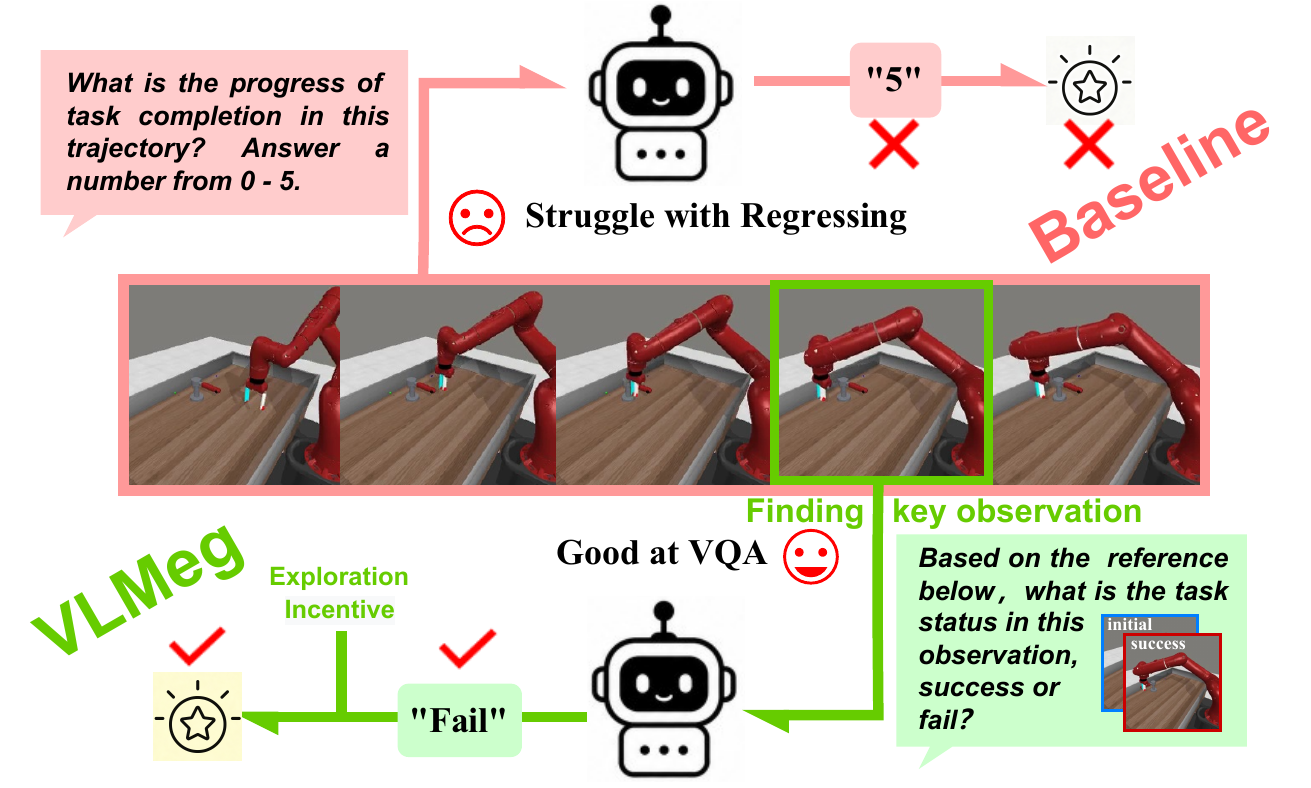}
\caption{Regressing precise rewards is misaligned with what VLMs naturally excel at. By contrast, serving as success detectors is more suitable for VLMs because it fully leverages their innate visual-question answering (VQA) capabilities. 
VLMeg is correspondingly proposed to well harness VLM in reward design, enabling more accurate embodied reward and more efficient RL.
}
\label{fig:detector-motivation}
\end{figure}

Reinforcement learning (RL) provides effective solutions for robotic tasks \cite{sutton2018rl,hilserl,comsd}, while designing effective rewards for diverse embodied tasks that involve different semantics and complex visual observation spaces remains another major challenge \cite{christiano2017preferences,eureka}.
With the rapid progress of large models, vision-language models (VLMs) have obtained increasingly strong general visual-language understanding ability and show promising generalization to embodied visual data and task semantics \cite{radford2021clip,alayrac2022flamingo}. Leveraging VLMs for embodied reward design has therefore become a promising and increasingly active direction \cite{successVQA,gptsuccess,rl-vlm-f,fu2024furl,venuto2024codeasreward,robofume}. 

A mainstream VLM-as-reward paradigm constructs reward-prediction datasets and trains VLMs to predict dense rewards \cite{roboreward,vlac,robodopamine,robometer,dino}. This paradigm not only requires extensive data annotation, but also prevents the use of leading closed-source VLMs. Moreover, although the learned dense reward models can perform well when fine-tuning pretrained policies like vision-language-action (VLA) models, they are not good at guiding RL from scratch, which should be a fundamental responsibility of a reward. This phenomenon stems from a basic mismatch: reward regression is not what VLMs are naturally designed for. We therefore propose that VLMs should be introduced into embodied RL reward design as success detectors (Figure~\ref{fig:detector-motivation}), so as to better exploit their innate visual-language reasoning and question-answering capability without additional post-training.

Recent works have explored and recognized the zero-shot ability of VLMs to judge robot task states \cite{successVQA,gptsuccess}. Nevertheless, how to organically use this ability in embodied RL rewards remains an open problem and faces several challenges. First, the inference cost and latency of large models are non-negligible, especially for leading closed-source models, while RL usually requires a massive number of environment samples. How to avoid redundant VLM usage and reduce the training burden is therefore critical. Second, although judging robot tasks is more aligned with VLM capability than directly regressing rewards, it is still challenging because it requires strong ability of 3D spatial reasoning and embodied semantic understanding. Compared with expert demonstrations, RL trajectories are more complex and diverse, containing many observations with occlusion or ambiguity that are even difficult for humans to judge. Inaccurate predictions with these low-quality observations may introduce inaccurate training signals that may hurt RL training. Third, a binary success-or-failure signal is extremely sparse, especially for early-stage policies. For RL from scratch, relying only on sparse success rewards often leads to poor sample efficiency and may even prevent policy convergence.

To address these challenges, we propose a novel embodied RL reward, dubbed \textbf{VLM success reward via Exploration-Augmented Key-Observation Gating (VLMeg)}, to better harness VLMs for embodied RL. Considering the diversity and complexity of embodied tasks, as well as the difficulty of 3D spatial judgment for VLMs, VLMeg provides the VLM with the task initial image and task success image as references, thereby reducing the difficulty of success detection. This design requires a success reference frame from expert videos. Considering the diversity of RL episodes, VLMeg finetunes a video generation model as episodic expert guide to substantially reduce the demand for expert videos, which follows recent works \cite{dreamgen,deg}. On this basis, VLMeg proposes an Exploration-Augmented Key-Observation Gating mechanism as the core interface for harnessing VLMs. During RL rollout, the gate compares each collected observation with the generated success reference frames and marks the episode as judgment-worthy once any observation satisfies the similarity threshold, thereby reducing redundant VLM calls. After the rollout ends, VLMeg selects the key observation with the highest similarity from all observations in the judgment-worthy trajectory as the representative for VLM judgment. This strategy avoids noisy observations such as occlusions and reduces visual distraction between the judged image and the success reference, thus improving VLM detection accuracy. Considering the sparsity of judgment-worthy trajectories, VLMeg further introduces state-of-the-art video-imitation exploration incentives based on the generated episodic guidance, which improves the efficiency of exploring toward judgment-worthy trajectories, triggers VLM rewards more frequently, and ultimately improves RL sample efficiency. Through the proposed Exploration-Augmented Key-Observation Gating, VLMeg organically integrates VLMs into embodied reward design as success detectors without VLM post-training, fully exploiting their native visual-language question-answering capability for robust and efficient embodied RL.

We conduct experiments on challenging reward-free embodied manipulation tasks in MetaWorld\cite{metaworld}. VLMeg achieves state-of-the-art from-scratch RL, outperforming advanced VLM-based approaches on all 12 tasks. We further conduct several experiments with a real-world Franka robotic arm. VLMeg enables effective RL without any ground-truth reward, outperforming baselines on all 3 real-world tasks.

Our contributions are summarized as follows:

\begin{itemize}
    \item Considering that regressing precise rewards is misaligned with what VLMs naturally excel at, we propose to employ VLMs as success detectors in embodied reward to fully leverage their innate multi-modal answering capabilities. We correspondingly propose VLMeg, a VLM-based embodied reward that harnesses frozen VLMs as episodic success detectors for efficient embodied RL.

    \item We design Exploration-Augmented Key-Observation Gating, which performs rollout-time trajectory filtering with generated expert videos and post-rollout key-observation selection from the whole judgment-worthy trajectory, reducing redundant VLM calls while avoiding occlusion and visual distraction. In addition, an advanced video-imitation reward is introduced, to facilitate exploration towards sparse judgment-worthy trajectories.
    

    \item We conduct extensive experiments across both simulation and real-world platforms. VLMeg achieves state-of-the-art RL without any ground-truth rewards, surpassing advanced VLM-based baselines on 12 MetaWorld tasks and 3 real-world Franka tasks. We also provide extensive analytical experiments to verify the effectiveness of each module and support our motivation. 
    

\end{itemize}

\section{Related Works}
\subsection{RL for Embodied Manipulation}
Robotic manipulation has advanced rapidly through imitation learning, which learns policies directly from expert demonstrations and avoids explicit reward design. However, behavior cloning remains restricted by demonstration coverage \cite{bcv-lr}: new situations cause small action errors to pile up, and suboptimal trajectories cannot be used naively because their actions are imitated as targets. Reward-learning variants such as adversarial imitation learning \cite{GAIL} and T-REX \cite{T-REX} alleviate this limitation by inferring training goals from demonstrations, but remain tied to offline data. Reinforcement learning can instead refine behavior through environmental interaction and improve beyond demonstrations\cite{hilserl}. However, embodied RL is frequently bottlenecked by reward engineering \cite{jingbo}. 

\subsection{RL Reward Design}
Reward functions in embodied RL are commonly categorized as sparse or dense. Sparse rewards provide a binary success signal and are widely used in real-world manipulation due to their simplicity and stability. For example, HIL-SERL employs a lightweight ResNet-based binary reward classifier to supervise robot learning \cite{hilserl}. However, because no intermediate feedback is available before task completion, sparse rewards often suffer from poor sample efficiency. To address this limitation, dense rewards estimate task progress and provide continuous guidance during training. Existing methods infer dense rewards through adversarial learning \cite{laifo}, visual representation learning \cite{vip,liv}, or generative video models \cite{tevir,deg}. Although these approaches substantially improve exploration and learning efficiency, inaccurate progress estimation can mislead the policy toward visually plausible yet unsuccessful states. 
\subsection{VLM for Embodied Reward}
The visual and linguistic knowledge of vision-language models (VLMs) has recently made them attractive reward interfaces for embodied RL. Early works mainly employ VLMs for sparse success detection by fine-tuning them to classify task completion \cite{successVQA,gptsuccess}, and later integrate these binary predictions into RL training \cite{robofume}. More recent approaches extend VLMs beyond binary supervision by learning dense or process-aware rewards, including progress prediction \cite{vlac}, process-aware reward modeling \cite{robodopamine}, video-language reward models \cite{roboreward}, and joint progress-success prediction \cite{robometer}. These methods significantly improve reward quality and generalization, but typically rely on large robot datasets, extensive negative-data construction, or dedicated reward-model training. Our work takes a different perspective by treating the VLM as a frozen success verifier rather than a trainable reward model, which fully leverages the innate multi-modal question-answering capabilities of VLM and achieves better performance on challenging from-scratch RL.


\section{Methodology}


VLMeg obtains episode-conditioned success reference frames from expert trajectories generated by a finetuned video generation model, and then harnesses VLM in embodied reward via the proposed Exploration-Augmented Key-Observation Gating. The gating mechanism first filters out low-quality RL episodes whose observations never approach the success reference region. Then, for each retained episode, it selects the most representative key observation and queries the VLM for an episodic success reward. Finally, it adds lightweight exploration incentives to address the sparsity of judgment-worthy episodes. The complete pipeline is shown in Figure \ref{fig:framework} and we provide the detailed pseudo code in Appendix B. 

\begin{figure*}[!t]
\centering
\includegraphics[width=1\textwidth]{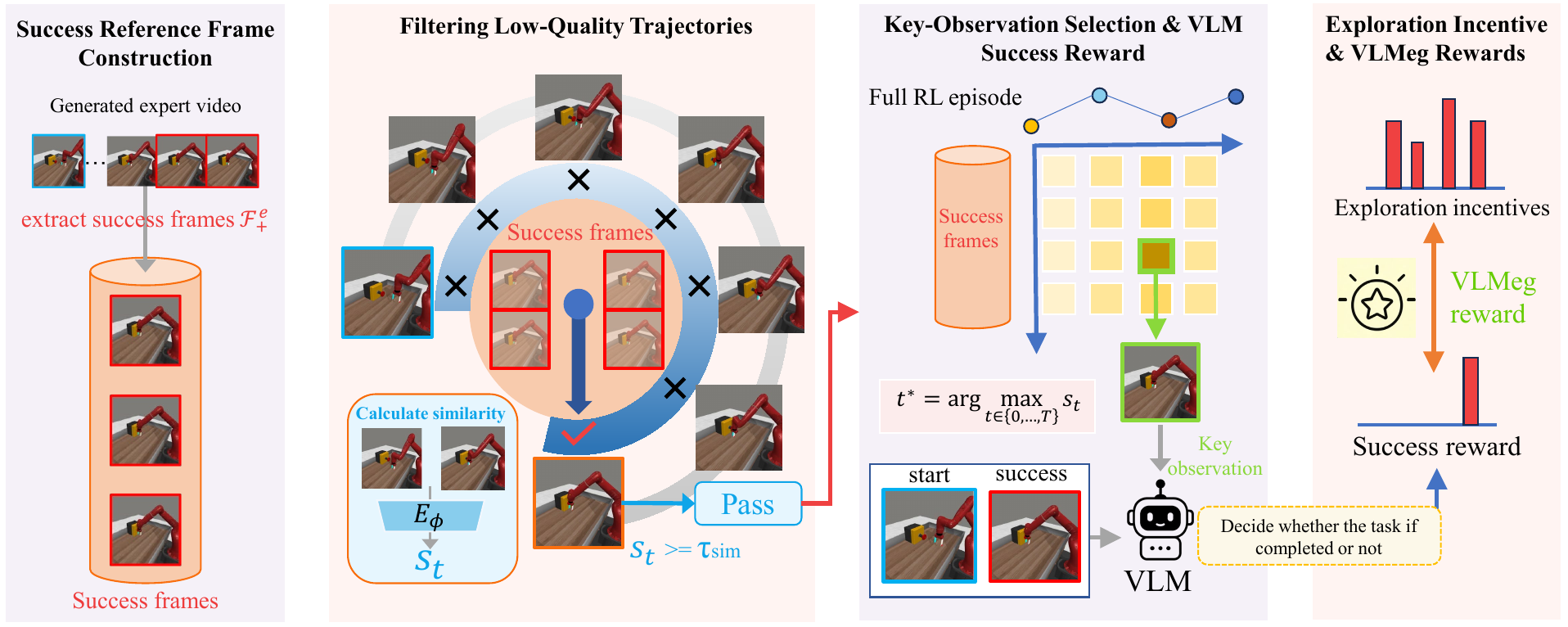}
\caption{Overview of VLMeg. First, successful frames are extracted from the generated expert videos. During rollout, each observation $o_t$ is compared with the success frames. If the maximum similarity exceeds the threshold $\tau_{\mathrm{sim}}$, the trajectory is considered worthy of VLM verification; otherwise, rollout continues until the horizon $T$, and trajectories that never satisfy the threshold skip VLM evaluation. For qualified trajectories, the observation most similar to the success-frame set is selected as the key observation. The initial frame, success reference, and selected key observation are then sent to the frozen VLM for success detection, and the resulting episodic success reward is combined with exploration incentives to improve RL efficiency.}
\label{fig:framework}
\end{figure*}

\subsection{Success Reference Frame Construction }
As shown in Table~\ref{tab:context-motivation}, judging task completion from only current observation is unreliable because the VLM lacks clear visual evidence of what the completed task state should look like. VLMeg therefore adopts a comparison-oriented success-detection architecture. Instead of asking the VLM to infer task completion from an isolated observation, VLMeg provides a completed-state visual reference together with the initial frame and the current state. This input structure helps the VLM focus on the task-relevant object-state change and reduces ambiguity caused by diverse objects, camera views, and robot poses.

However, obtaining such references is nontrivial in online RL. Episodes can start from diverse initial states, while collecting expert videos that cover these variations is expensive. VLMeg therefore obtains episode-specific success references with a video generation model. We adapt an image-to-video foundation model $G$ to the target task with LoRA \cite{lora} using the few expert videos and the task instruction:

\begin{equation}
G_\psi = \mathrm{LoRA}(G;\mathcal{V},q).
\end{equation}

At the beginning of episode $e$, the adapted generator receives the actual initial observation $o_0^e$ and the task prompt $q$, and generates an expert-like successful trajectory:

\begin{equation}
\tilde{V}^e = G_\psi(o_0^e,q), \qquad
\tilde{V}^e=(\tilde{v}_1^e,\ldots,\tilde{v}_l^e).
\end{equation}

Since the generated trajectory is expected to complete the task, its later frames provide success-state evidence. We therefore keep the frames after an index threshold $\tau_{\mathrm{idx}}$ as the success reference frames:

\begin{equation}
\mathcal{I}_+^e=\{i\in\{1,\ldots,l\}\mid i\geq\tau_{\mathrm{idx}}\},
\qquad
\mathcal{F}_+^e=\{\tilde{v}_i^e\mid i\in\mathcal{I}_+^e\}.
\end{equation}

\begin{table}[!t]
\centering
\caption{Motivation study for context-aware VLM success detection on \texttt{faucet-close}. The dataset used is described in Appendix C.}
\label{tab:context-motivation}
\begin{tabular}{lc}
\toprule
VLM input & Acc. \\
\midrule
w/ Reference Frames & 100\% \\
w/o Reference Frames & 42\% \\
\bottomrule
\end{tabular}
\end{table}

\subsection{Filtering Low-Quality Trajectories}
Most episodes during early from-scratch RL are far from success, and calling a VLM on them is both expensive and unreliable. VLMeg therefore first filters out low-quality trajectories before any VLM query. This filtering is performed online during rollout and is used only to decide whether the whole trajectory is worth VLM judgment. It relies on a task-specific similarity encoder $E_\phi$. To keep the encoder lightweight and aligned with the task domain, we train it from the expert videos using a contrastive self-supervised objective: two augmented views of the same expert frame are treated as positives, while different frames in the mini-batch are treated as negatives. The trained encoder maps expert frames, generated reference frames, and online observations into a shared latent space.

Given the success reference frames $\mathcal{F}_+^e$, every time a new observation $o_t$ is collected during rollout, VLMeg compares it with the retained success frames and computes the maximum cosine similarity:

\begin{equation}
s_t = \max_{i \in \mathcal{I}_+^e}
\cos(E_\phi(o_t),E_\phi(\tilde{v}_i^e)).
\end{equation}

The similarity score $s_t$ measures whether observation $o_t$ is close enough to the generated success region. Because $s_t$ is computed as observations arrive, VLMeg can determine during rollout whether the trajectory is worth VLM judgment: once any frame reaches the similarity threshold $\tau_{\mathrm{sim}}$, the current episode is regarded as judgment-worthy. VLMeg does not immediately query the VLM when the threshold is triggered; instead, the rollout continues to completion and then enters the key-observation selection stage. If the rollout ends without any frame reaching the threshold, the trajectory is discarded for VLM evaluation, the VLM call is skipped, and the episodic success reward is set to zero. This gate removes redundant evaluations and prevents the VLM from being forced to judge unsupported exploratory trajectories.

\subsection{Key-Observation Selection \& VLM Success Reward}
For a trajectory regarded as judgment-worthy during rollout, VLMeg performs key-observation selection after the episode rollout is complete. Although the trajectory contains at least one observation close to the success reference frames, it may still include many unsuitable frames. Some frames are occluded by the robot arm, some contain distracting intermediate poses, and some have already passed the clearest completion moment. VLMeg therefore needs to select one representative key observation after the rollout ends.

\begin{table}[!t]
\centering
\caption{We construct two datasets: Deviated and Near. Deviated contains images where the robot arm states significantly deviate from the expert frames, while Near contains images where the robot arm states are close to those in the expert frames. More details about the datasets are described in Appendix C.}
\label{tab:key-observation-motivation}
\begin{tabular}{lcc}
\toprule
Dataset&Acc.(successful data)&Acc.(failing data)\\
\midrule
Deviated& 20\% & 80\% \\
Near& 100\% & 100\% \\
\bottomrule
\end{tabular}
\end{table}

The diagnostic result in Table~\ref{tab:key-observation-motivation} provides the motivation for this selection rule on \texttt{faucet-close}. Under the same unobstructed setting, the VLM struggles on Deviated dataset  but reliably identifies Near set. Therefore, the key observation should be the frame closest to the generated success frames in the learned similarity space. VLMeg implements this principle by selecting the maximum-similarity observation from the whole judgment-worthy trajectory:

\begin{equation}
t^\star=\arg\max_{t\in\{0,\ldots,T\}}s_t.
\end{equation}

The selected observation $o_{t^\star}$ represents the whole episode for VLM judgment. This strategy avoids blindly using the terminal frame, which may be occluded by the robot arm or may miss the clearest completion moment. It also reduces visual distraction because frames far from the generated success frames are less likely to be selected. The visual reference passed to the VLM is fixed as the final frame of the generated expert trajectory:

\begin{equation}
v_{\mathrm{ref}}^e=\tilde{v}_{l}^e.
\end{equation}

The frozen VLM detector $D$ then receives the task prompt $q$, the initial frame $o_0$, the key observation $o_{t^\star}$, and the success reference frame $v_{\mathrm{ref}}^e$. The prompt asks the VLM to output $1$ if the task has been completed in the key observation relative to the initial frame and the success reference, and $0$ otherwise:

\begin{equation}
b(\tau)=D(q,o_0,o_{t^\star},v_{\mathrm{ref}}^e)\in\{0,1\}.
\end{equation}

Combining the low-quality trajectory filter and the VLM judgment, the episodic success reward is

\begin{equation}
R_{\mathrm{succ}}(\tau)=
\left\{
\begin{array}{ll}
b(\tau), & \mbox{judgment-worthy},\\
0, & \mbox{discarded}.
\end{array}
\right.
\end{equation}

\begin{figure*}[t]
\centering
\includegraphics[width=0.98\textwidth]{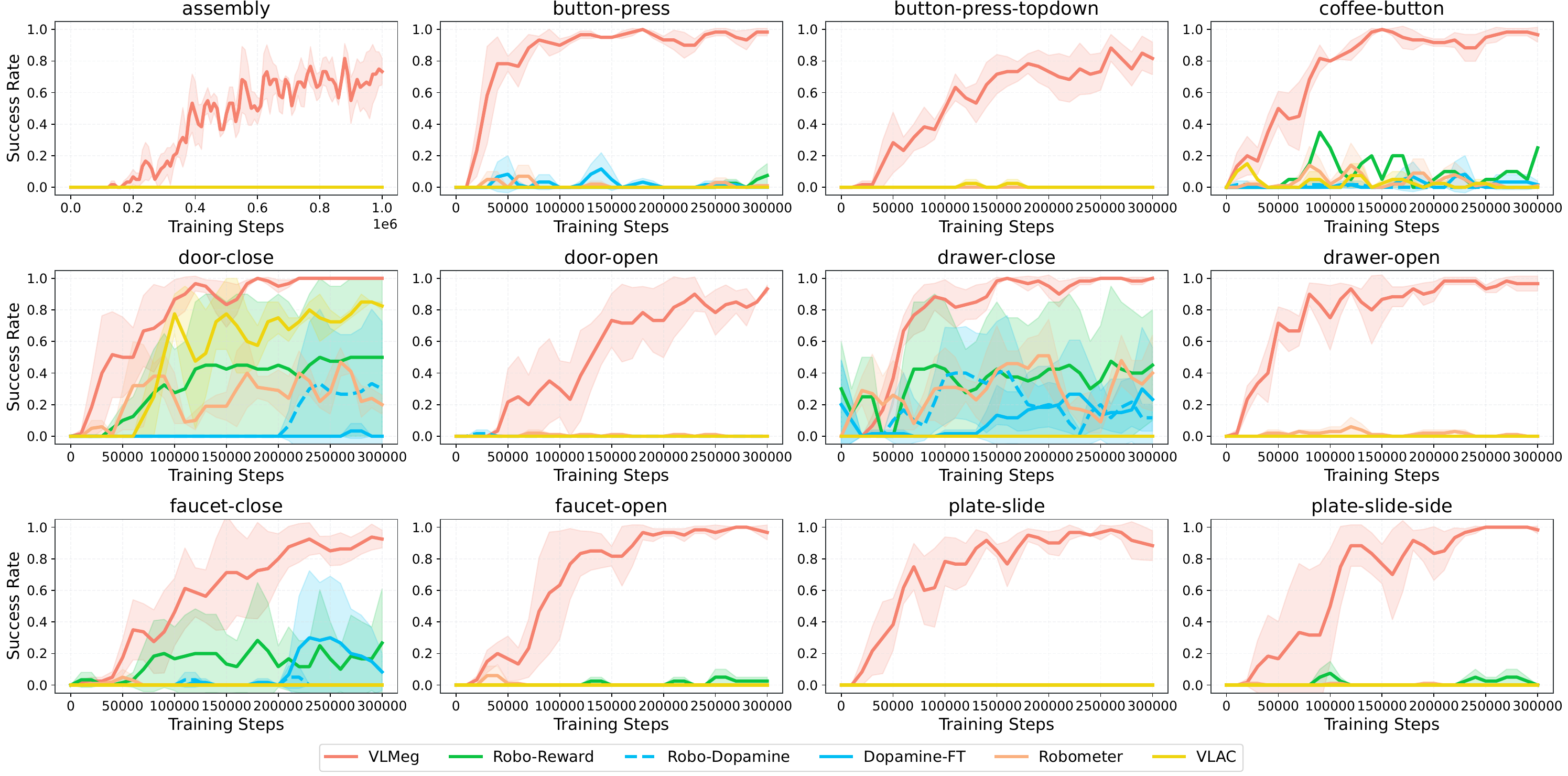}
\caption{Simulation results on the 12 MetaWorld reward-free manipulation tasks. VLMeg is compared with representative VLM-based reward baselines under from-scratch RL using the same task prompts and expert-video budget.}
\label{fig:sim-main}
\end{figure*}

This mechanism changes the role of the VLM. The VLM is not asked to score every state or regress a dense reward, but is invoked once for a carefully selected key observation from a judgment-worthy episode. The success reference frame, initial frame, and task prompt make the query easier for the VLM to understand and more aligned with its natural visual question-answering capability.

\subsection{Exploration Incentives \& VLMeg Reward}
Although the gate improves the efficiency and accuracy of VLM success detection, judgment-worthy episodes are sparse during early-stage from-scratch RL. VLMeg therefore adds exploration assistance based on the same generated trajectory. This term is used only to guide the agent toward states where the gate can be triggered more frequently; semantic task completion is still verified by the VLM success reward. We adopt a recent advanced video-imitation dense incentive \cite{deg} as a lightweight exploration term $r_{\mathrm{expl},t}$. It compares online observations with the generated guidance trajectory in the contrastive feature space, encouraging both coarse sequential progress along the guidance and fine-grained visual matching to task-relevant frames. We provide the computation details of this exploration incentive in Appendix B. 

VLMeg broadcasts the episodic VLM success reward to every timestep and combines it with the exploration incentive, obtaining the final VLMeg reward as follows:

\begin{equation}
r_t^{\mathrm{VLMeg}}=
r_{\mathrm{expl},t}+\lambda R_{\mathrm{succ}}(\tau), \qquad t=0,\ldots,T.
\end{equation}

\begin{figure}[!t]
\centering
\includegraphics[width=0.99\linewidth]{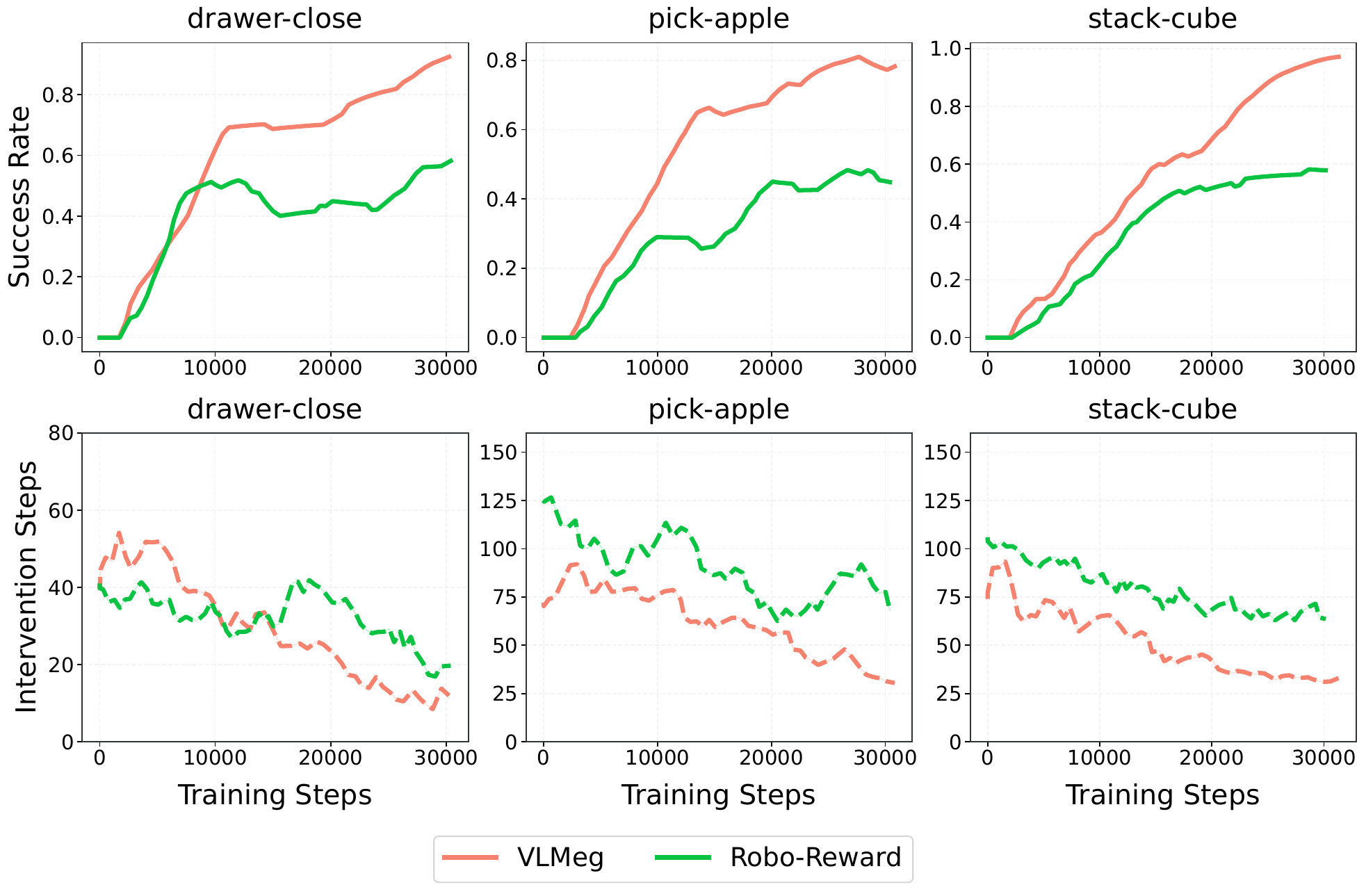}
\caption{Smoothed training curves on real-world Franka manipulation tasks. VLMeg enables better RL efficiency and lower intervention rates than Robo-Reward, demonstrating its effectiveness in physical manipulation RL.}
\label{fig:real-world}
\end{figure}

\section{Experiments}
\subsection{Experimental Setup}
\subsubsection{Tasks}

We evaluate VLMeg on 15 reward-free manipulation tasks from both simulation and real-world platforms. For simulation, we employ 12 challenging MetaWorld tasks.
For real-world evaluation, we use a Franka robotic arm and design 3 challenging long-horizon tasks. More details about environments and tasks are provided in Appendix C.
\subsubsection{Baselines}
We compare VLMeg with five state-of-the-art VLM-based embodied reward baselines: Robo-Reward \cite{roboreward}, Robometer \cite{robometer}, VLAC \cite{vlac}, Robo-Dopamine \cite{robodopamine}, and Dopamine-FT \cite{robodopamine}. 

Robo-Reward and Robometer are trajectory-level reward models. They take a downsampled episode as input and output a task-progress reward for the whole trajectory. 
VLAC and Robo-Dopamine are frame-pair reward models. They fine-tune a model to estimate the relative progress between two frames and then calculate rewards by accumulating the predicted progress across frame pairs. Robo-Dopamine can be fine-tuned using expert data for specific tasks. Therefore, we use the same expert data used in VLMeg for reward annotation and model fine-tuning, referring to this variant as Dopamine-FT. Refer to Appendix C for detailed baseline introductions.

\subsubsection{Implementation of VLMeg}
The video generator used by VLMeg is Wan2.1-I2V-14B \cite{wan}. For each task, we provide 3--5 expert videos and adapt the generator with LoRA \cite{lora}. 
The contrastive encoder is trained for 5k updates with a learning rate $1\times10^{-4}$. 
The weight of the episodic VLM success reward is set to $\lambda=10$. 
The similarity threshold is fixed as $\tau_{\mathrm{sim}}=0.95$ for all experiments. All VLM calls in VLMeg are made with GPT-5.5. Refer to Appendix C for detailed hyper-parameter settings of VLMeg.

\subsection{VLMeg on Simulation Tasks}
\subsubsection{Comparison with Baselines}
To evaluate the practical effectiveness of different reward models in embodied manipulation, we use VLMeg and all baselines as RL rewards on the 12 MetaWorld tasks. DrQv2 is used as the default RL backbone. For Robometer, we instead use SAC to align with its officially released RL implementation. All methods are trained for 300k environment steps, except for \texttt{assembly}, which is trained for 1000k steps due to its difficulty. We only provide the task prompt for VLM inference and 3–5 single-view expert trajectories for fine-tuning both the VLMeg video generation model and Dopamine-FT. No environment reward is used. We provide detailed settings of VLMeg and baselines in Appendix C.

Figure~\ref{fig:sim-main} reports the main results on the 12 MetaWorld tasks. VLMeg consistently achieves stronger from-scratch RL performance than the VLM-reward baselines. The advantage is most evident on tasks requiring precise object interaction, where visual progress estimation can be difficult. However, on tasks with lower precision requirements, such as \texttt{door-close} and \texttt{drawer-close}, several baselines can obtain relatively high success rates. Dopamine-FT also improves over the non-fine-tuned Robo-Dopamine on some tasks, showing the benefit of task-specific data. However, its success rate remains much lower than VLMeg, and it can even underperform the original Robo-Dopamine on certain tasks such as \texttt{door-close}. This suggests that 3--5 similar successful trajectories are still insufficient for fine-tuning a dense VLM reward model.

\subsubsection{Computational Consumption}

\begin{figure*}[t]
\centering
\includegraphics[width=0.98\linewidth]{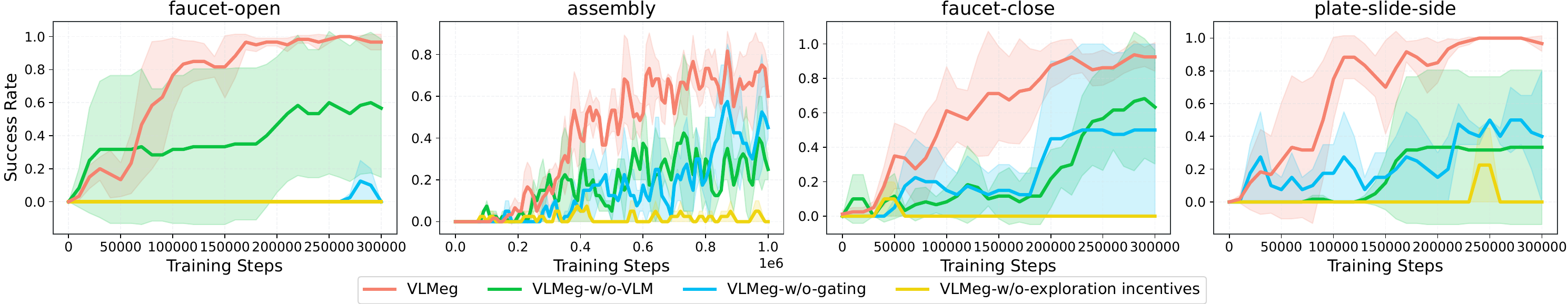}
\caption{Ablation results on four MetaWorld tasks. The comparison isolates the contribution of dense incentives, gated key-observation selection, and episodic VLM success verification.}
\label{fig:ablation}
\end{figure*}

\begin{figure}[!t]
\centering
\includegraphics[width=0.98\linewidth]{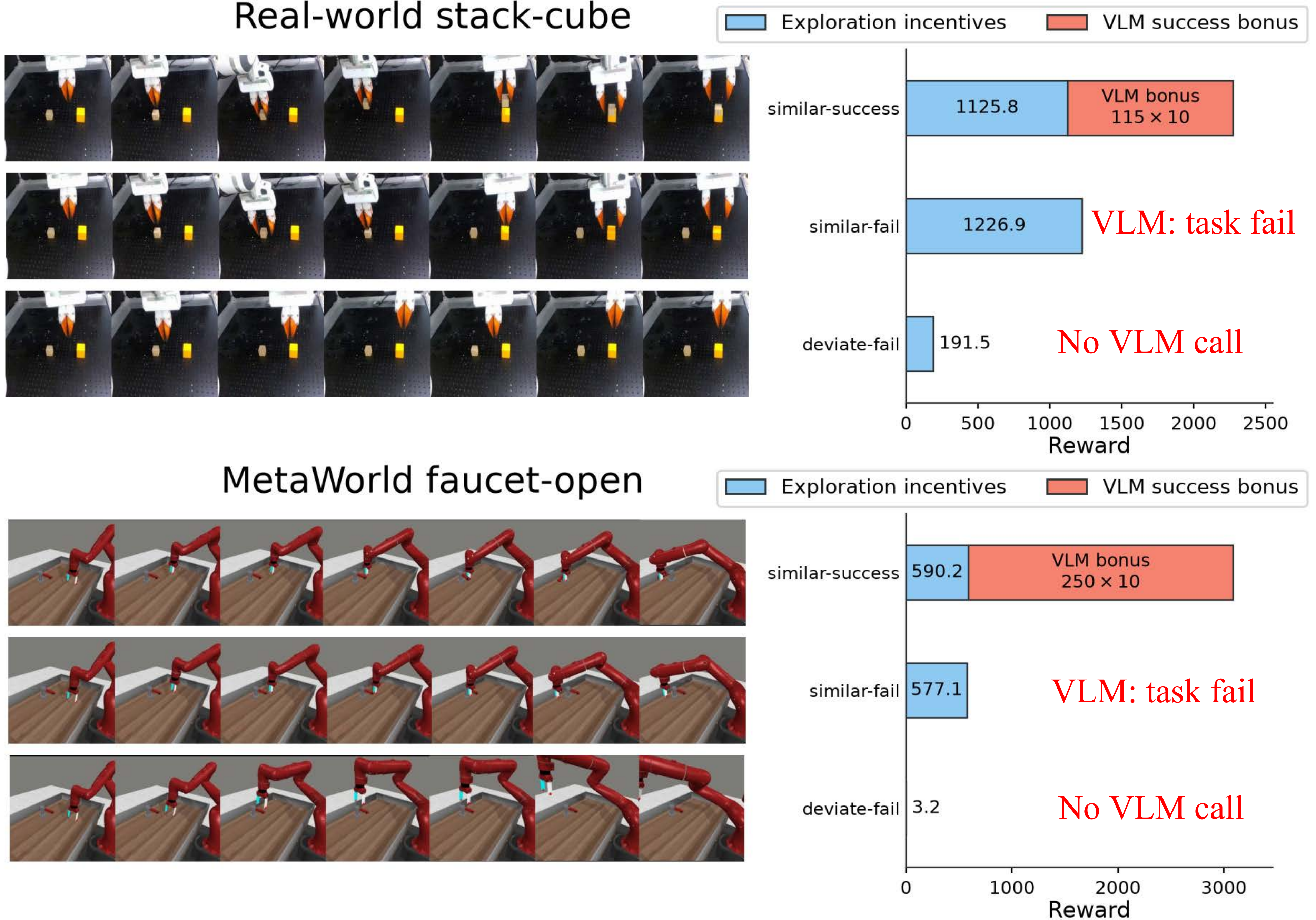}
\caption{Comparison of cumulative exploration incentives and VLM rewards obtained by different trajectories.}
\label{fig:reward-vis}
\end{figure}

We further compare the computational consumption of different reward designs. The video generation process in VLMeg takes about 3.1 hours, while the remaining training process requires about 3.5 hours, resulting in a total training time of 6.6 hours for 300k RL steps. This is slightly higher than Robo-Dopamine and Robo-Reward, which require 5.5 and 5.2 hours, respectively. Despite the additional computation cost, VLMeg achieves substantially better policy learning performance, demonstrating the effectiveness of its video-based guidance and VLM verification.

\subsection{Real-World Experiments}

We further evaluate VLMeg on three physical manipulation tasks with a Franka robotic arm: \texttt{drawer-close}, \texttt{pick-apple}, and \texttt{stack-cube}. Besides VLMeg, we also deploy Robo-Reward
\cite{roboreward} as the real-world baseline. Both methods use only a single-view camera observation, a language instruction, and three successful expert videos collected from different initial object positions. The expert videos are used for video-generator adaptation in VLMeg. We use HIL-SERL \cite{hilserl} as the real-world RL backbone and train each task for 30k steps. Figure~\ref{fig:real-world} reports the real-world learning behavior.

During training, we observe that human-in-the-loop guidance helps the policy reach states that are more similar to the generated expert guidance, which makes the episodic success reward easier to trigger. The VLMeg success reward can replace the human-provided binary keyboard reward, substantially reducing the human labeling burden during real-world RL. With VLMeg, human interventions decrease gradually as training proceeds, while success rates rise steadily, demonstrating effective policy learning. In contrast, Robo-Reward struggles to distinguish near-success trajectories from truly successful trajectories. As a result, the reward-free policy is difficult to optimize, human interventions remain frequent, and the success rate fails to improve reliably.

\subsection{Ablation Study}

Following the simulation setup above, we conduct ablation experiments on four MetaWorld tasks. Figure~\ref{fig:ablation} shows the results. \textbf{VLMeg-w/o-VLM} removes the VLM success reward and uses only the exploration incentives. This variant can provide efficient exploration, but it may assign a high reward to trajectories that visually imitate the expert arm motion without completing the object interaction. \textbf{VLMeg-w/o-gating} keeps the exploration incentives and queries the VLM on the fixed final observation of each episode. This variant tests whether a naive terminal-frame VLM detector is enough. Its weaker performance shows that the final frame is not always the most informative observation. \textbf{VLMeg-w/o-exploration incentives} removes the exploration incentives and uses only the gating mechanism with VLM success detection. This variant provides more reliable success labels than ungated VLM querying, but the sparse signal alone is insufficient for sample-efficient exploration. In contrast, VLMeg effectively combines the strengths of all the components, enabling efficient exploration and accurate task completion. We provide more ablation and additional hyper-parameter sensitivity experiments. Refer to Appendix A for details.

\subsection{VLMeg Reward on Different Trajectories}

To better understand why episodic VLM verification is necessary, we compare representative trajectories on both a simulation task and a real-world task in Figure~\ref{fig:reward-vis}. For trajectories that largely deviate from expert behaviors (\textbf{deviate-fail}), exploration incentives can already distinguish them from high-value trajectories that are more likely to succeed (\textbf{similar-success} \& \textbf{similar-fail}). Meanwhile, our gating mechanism effectively filters out such trajectories without invoking VLM verification, reducing unnecessary inference costs.

In contrast, for trajectories with highly similar robot motions but subtle differences in object states that determine success (\textbf{similar-success}) or failure (\textbf{similar-fail}), exploration incentives struggle to provide discriminative rewards. In these cases, the gating mechanism identifies both trajectories as worthy of further evaluation and activates the VLM success detector. By leveraging semantic understanding, VLM provides accurate success rewards, enabling more accurate policy optimization.

\begin{figure}[!t]
\centering
\includegraphics[width=0.96\linewidth]{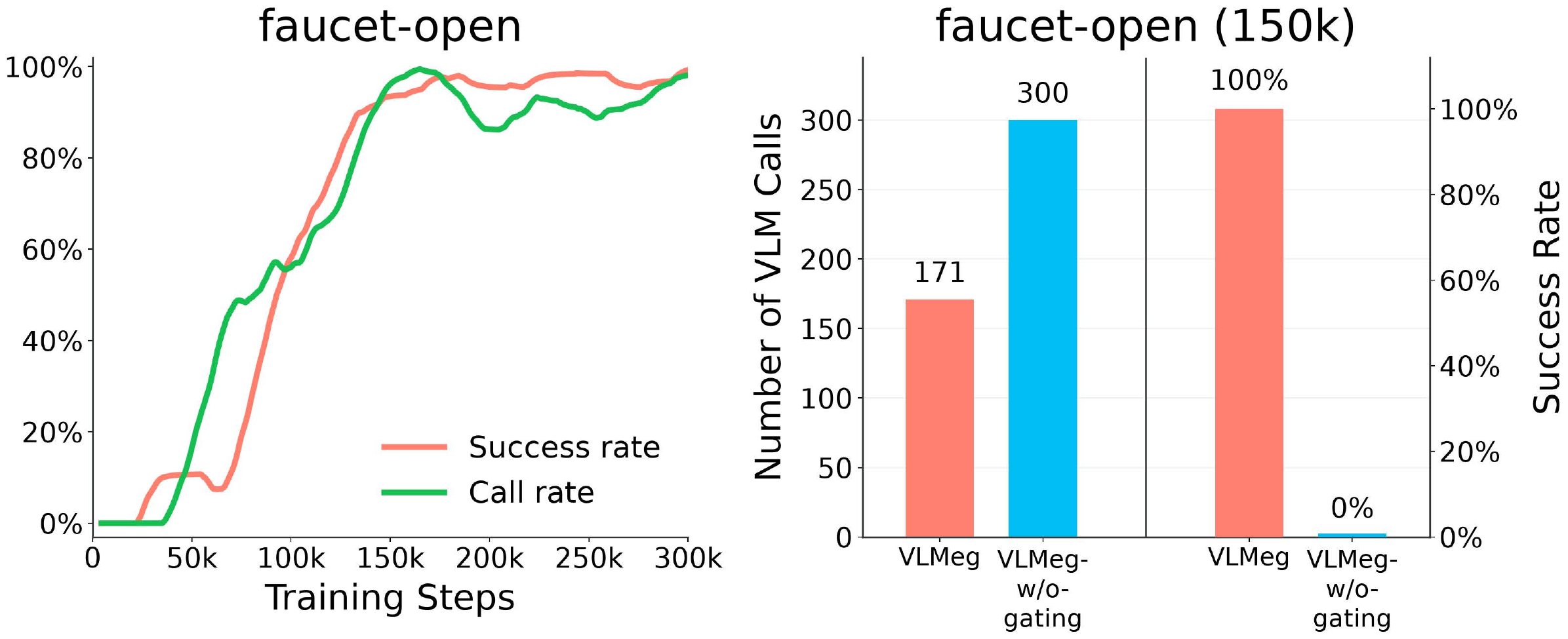}
\caption{The left figure shows task success rate and VLM call rate during training. The right figure shows the number of VLM calls and task success rate before policy convergence (150k steps). VLMeg is our proposed method, while VLMeg-w/o-gating directly uses the last frame of the trajectory as the key observation.}
\label{fig:call-rate}
\end{figure}

\subsection{Effectiveness of Key-Observation Gating}
As shown in Figure~\ref{fig:call-rate}, we evaluate the effectiveness of the key-observation gating mechanism from both call efficiency and learning performance. The left curve shows that the VLM call rate is highly correlated with the task success rate. 
During early training, the gating mechanism reduces unnecessary VLM calls when trajectory quality is low. As the policy improves, the call rate increases accordingly. The right comparison further demonstrates the effectiveness of the gating mechanism. Compared with VLMeg w/o gating, VLMeg achieves better policy learning performance with fewer VLM calls and less inference time. This verifies that the gating mechanism not only alleviates the computational burden of VLM usage but also improves the quality of VLM calls by selectively focusing on informative trajectories. More results on judgment accuracy are provided in Appendix A.
\section{Conclusion}
This paper revisits the role of VLMs in embodied reward design and argues that success detection is preferable to dense reward regression. VLMeg realizes this view through exploration-augmented key-observation gating, which filters low-quality trajectories, selects informative observations, and queries a frozen VLM only when needed. Across simulation and real-world tasks, VLMeg improves reward accuracy, reduces VLM usage, and enables efficient reward-free RL. These results suggest a practical direction for integrating foundation models into embodied policy learning.

\bibliography{aaai2027}

\newpage
\onecolumn

{\huge \textbf{Appendix}}

\section{A. Additional Experimental Results}
\subsection{A.1. More Ablation Results}

\begin{figure}[h]
\centering
\includegraphics[width=0.8\linewidth]{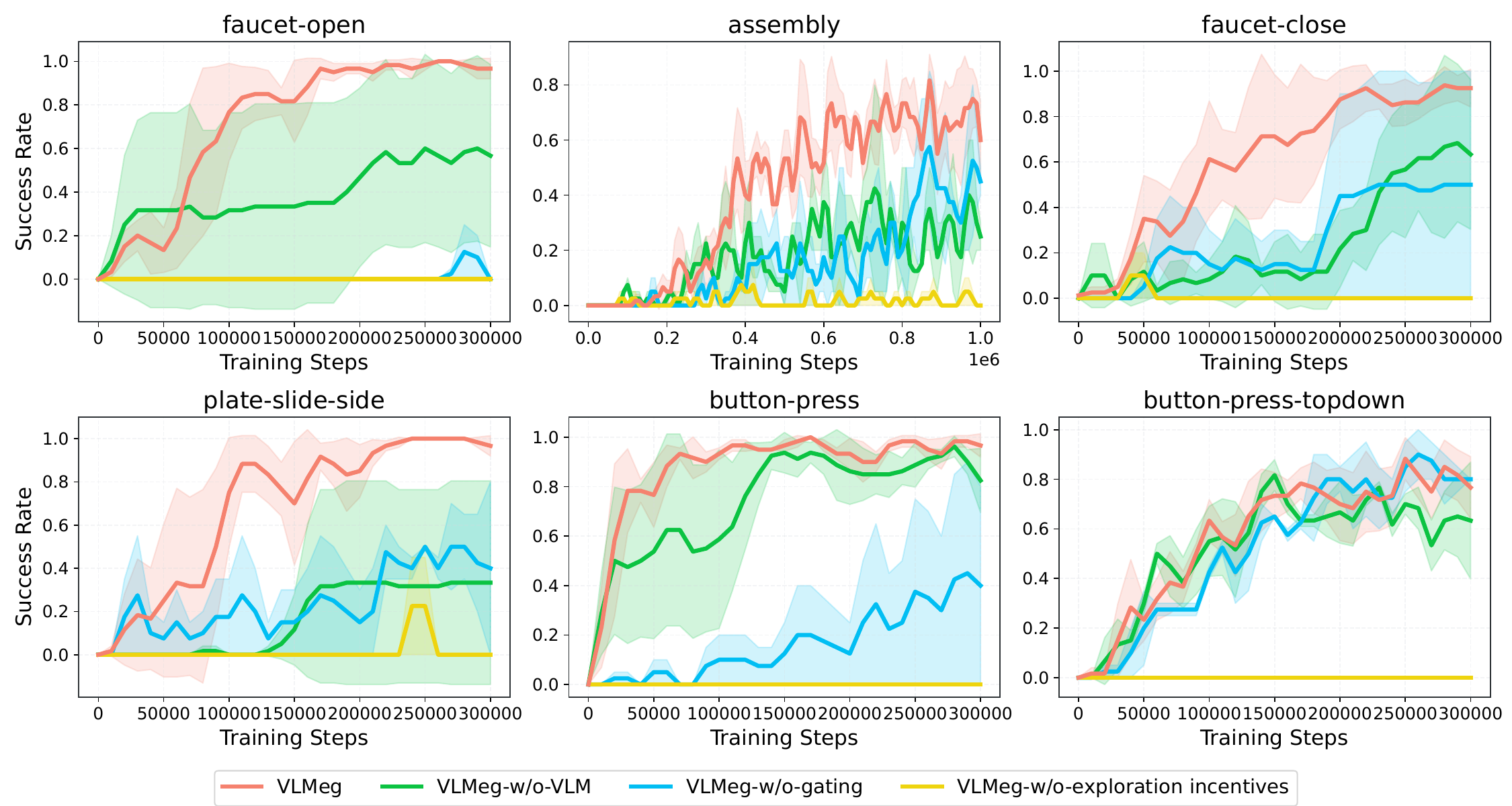}
\caption{Ablation results on six MetaWorld tasks. The comparison isolates the contribution of key-observation Gating, episodic VLM success verification, and exploration incentives.}
\label{fig:ablation6}
\end{figure}

In the main paper, we report ablation results on four MetaWorld tasks to analyze the contribution of the major components in VLMeg. In this section, we further extend the ablation study to six MetaWorld tasks, as shown in Figure~\ref{fig:ablation6}. The notation \textbf{VLMeg-w/o-[module]} denotes an ablated variant where the corresponding component is removed from the full method. Specifically, \textbf{VLMeg-w/o-VLM} removes the episodic VLM success reward and relies only on exploration incentives; \textbf{VLMeg-w/o-gating} disables the exploration-augmented key-observation gating and directly queries the VLM with the terminal observation of each episode; and \textbf{VLMeg-w/o-exploration incentives} removes the dense exploration term and keeps only gated VLM success detection. The results further verify that VLM success verification is needed to distinguish visually similar success and failure trajectories, gating is needed to reduce noisy or redundant VLM judgments and select informative observations, and exploration incentives are needed to make sparse success detection sample-efficient.

\clearpage
\subsection{A.2. Hyper-parameter sensitivity} 
\begin{figure}[h]
\centering
\includegraphics[width=0.5\linewidth]{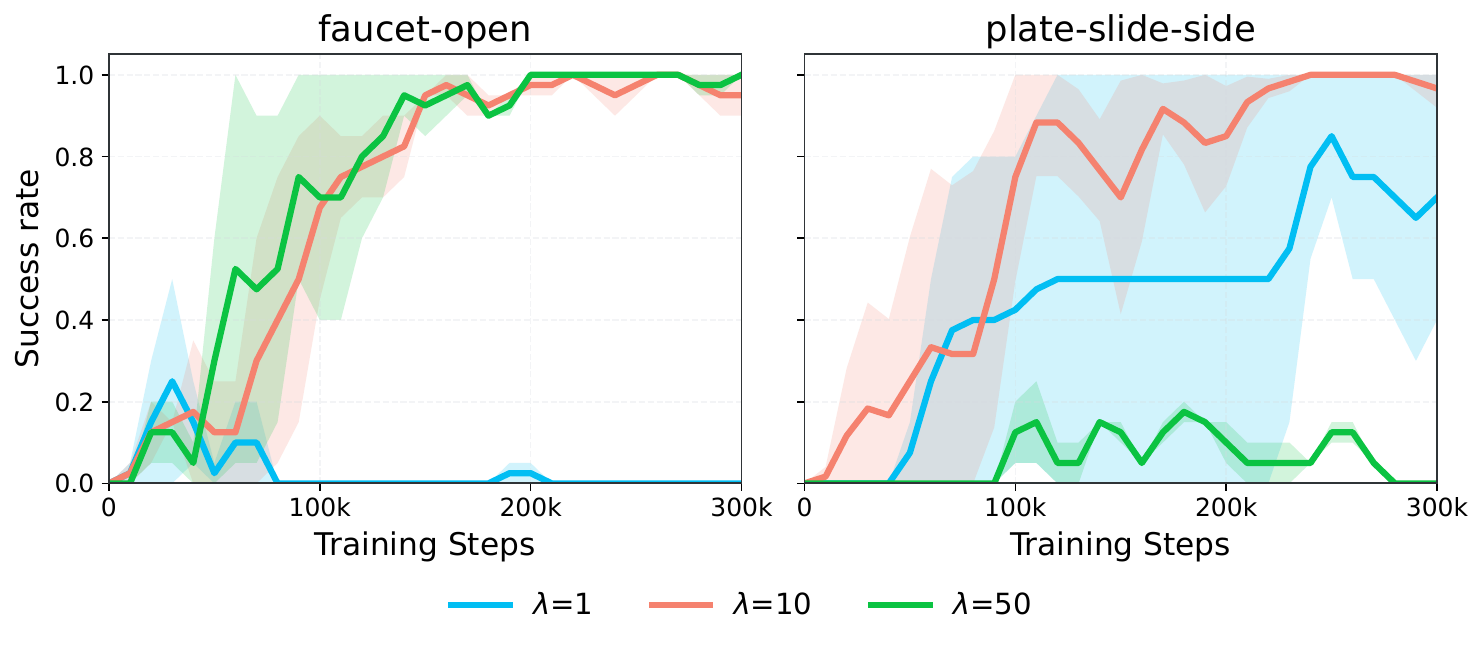}
\caption{Hyper-parameter sensitivity of the success-reward weight $\lambda$ on two MetaWorld tasks. The curves compare training success rates with $\lambda=1$, $\lambda=10$, and $\lambda=50$. A small $\lambda$ weakens the episodic VLM success signal, while an overly large $\lambda$ may overemphasize sparse success verification and reduce robustness on some tasks. The setting $\lambda=10$ provides the most consistent performance across the evaluated tasks.}
\label{fig:hyp}
\end{figure}

We further study the sensitivity of VLMeg to the success reward weight $\lambda$ in the VLMeg reward. Figure~\ref{fig:hyp} compares $\lambda=1$, $\lambda=10$, and $\lambda=50$ on \texttt{faucet-open} and \texttt{plate-slide-side}. When $\lambda$ is too small, the sparse VLM success reward is underweighted and the policy can be dominated by the dense exploration incentive, which leads to poor learning on \texttt{faucet-open} and slower convergence on \texttt{plate-slide-side}. Increasing $\lambda$ strengthens the influence of semantic success verification. However, an overly large weight does not consistently improve performance: although $\lambda=50$ performs well on \texttt{faucet-open}, it becomes unstable and substantially worse on \texttt{plate-slide-side}. In contrast, $\lambda=10$ achieves strong and stable performance on both tasks, suggesting that a moderate balance between dense exploration guidance and episodic VLM success verification is important for robust RL training. Therefore, we set $\lambda=10$ for all experiments in this paper.

\subsection{A.3. Comparison between Different VLM backbones in VLMeg} 

\begin{figure}[h]
\centering
\includegraphics[width=0.98\linewidth]{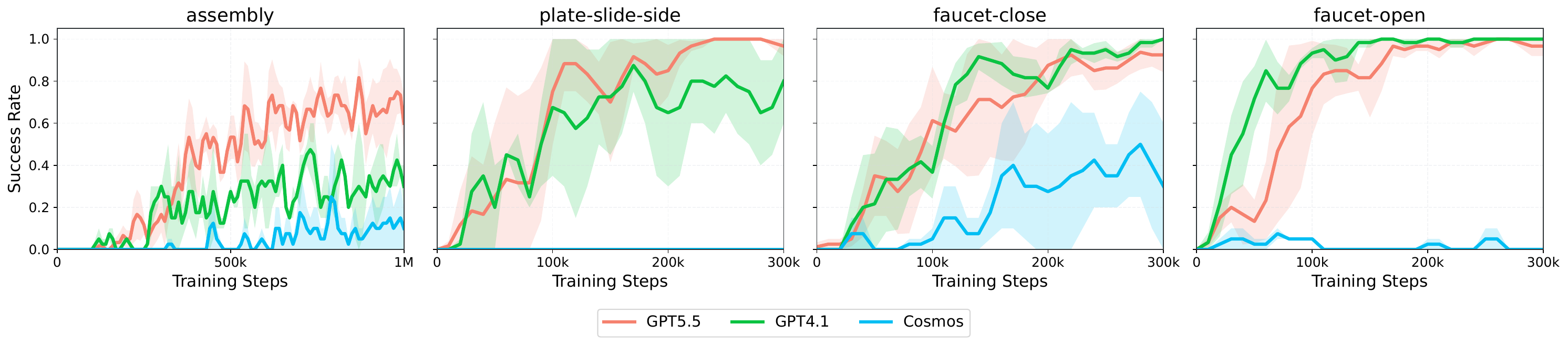}
\caption{Comparison between different VLM backbones in VLMeg on four MetaWorld tasks. GPT-5.5 and GPT-4.1 represent closed-source large VLMs, while Cosmos denotes the open-source smaller model Cosmos-Reason2-8B. The closed-source large VLMs provide substantially stronger success-detection signals than Cosmos-Reason2-8B, leading to better RL performance. GPT-5.5 is generally more stable and robust across tasks, while GPT-4.1 can also achieve strong performance on several tasks.}
\label{fig:dif-model}
\end{figure}

We compare different VLM backbones in VLMeg to study how the capability of the success detector affects downstream RL performance. As shown in Figure~\ref{fig:dif-model}, the closed-source large VLMs, GPT-5.5 and GPT-4.1, substantially outperform the open-source smaller model Cosmos-Reason2-8B\cite{nvidia2026cosmosreason2} on all the tested tasks. However, directly applying closed-source large VLMs in online RL can introduce high API cost and long inference latency. VLMeg addresses this issue through exploration-augmented key-observation gating. Therefore, VLMeg can benefit from the stronger reasoning ability of closed-source large VLMs while keeping their usage efficient. In all experiments of this paper, we use GPT-5.5 as the default VLM backbone. Meanwhile, GPT-4.1 also achieves competitive performance on several tasks, such as \texttt{faucet-close} and \texttt{faucet-open}, but it is less stable and generalizes less consistently than GPT-5.5 on some other tasks such as \texttt{assembly} and \texttt{plate-slide-side}.

\clearpage
\subsection{A.4. The Effectiveness of Key-Observation Gating: Calling times, Judgment Accuracy, and RL Performance}
\label{sec:gating-effectiveness}

In the ``Gating Mechanism for Efficient VLM Usage'' section of the main paper, we compare the VLM calling rate between VLMeg and VLMeg-w/o-gating. In this section, we provide additional results and further analyze how the proposed key-observation gating mechanism improves VLM detection accuracy. 

\begin{figure}[h]
\centering
\includegraphics[width=0.95\linewidth]{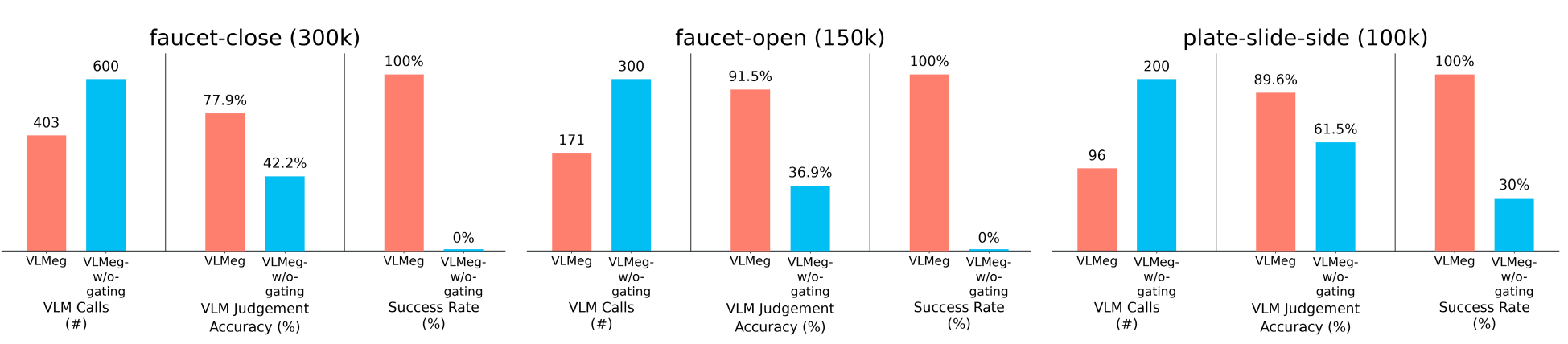}
\caption{Effectiveness of key-observation gating on \texttt{faucet-close}, \texttt{faucet-open}, and \texttt{plate-slide-side}. The number of training steps shown in each task title denotes the point at which VLMeg converges on that task; all VLM calls, VLM judgment accuracy, and task success rates are accumulated or evaluated up to this convergence point.
}
\label{fig:gating-effectiveness}
\end{figure}

Figure~\ref{fig:gating-effectiveness} demonstrates that key-observation gating improves VLM usage quality rather than merely reducing the number of queries. Across all tasks, VLMeg consistently achieves higher VLM judgment accuracy and task success rates with fewer VLM calls compared with the ungated variant. These results suggest that directly querying the VLM on ungated terminal observations often introduces ambiguous or visually unsupported states, leading to unreliable success judgments. By filtering out low-quality trajectories and selecting more informative observations for verification, VLMeg obtains more accurate success signals and provides stronger guidance for RL training.

\clearpage
\section{B. Supplemented Details of VLMeg Methodology}
\subsection{B.1. Detailed Computation of Exploration Incentive}
VLMeg uses a DEG-style video-imitation reward \cite{deg} as a dense exploration incentive. This term is not intended to determine semantic task success. Instead, it provides step-wise guidance that encourages the policy to visit states close to the episode-conditioned generated guidance video, thereby increasing the chance of reaching judgment-worthy episodes that can trigger the sparse VLM success reward.

The employed exploration incentive constructs dense guidance by comparing online observations with frames in an expert-like guidance video in a learned visual feature space. Here we reuse the generated trajectory $\tilde{V}^e=(\tilde{v}_1^e,\ldots,\tilde{v}_l^e)$ and the task-specific encoder $E_\phi$ described in the main paper. For an RL observation $o_t^e$ at timestep $t$ and the $i$-th generated guidance frame $\tilde{v}_i^e$, we compute their cosine similarity as

\begin{equation}
S(t,i)=\cos(E_\phi(o_t^e),E_\phi(\tilde{v}_i^e)).
\end{equation}

The exploration incentive contains two complementary components. The first one is a coarse-grained progress reward, which encourages the agent to move forward along the generated guidance sequence. Let $I_{\mathrm{target}}$ denote the current guidance-frame index to be approached, initialized as $I_{\mathrm{target}}=1$ at the beginning of each episode. The coarse reward measures the similarity improvement toward this target frame:

\begin{equation}
r_{\mathrm{coarse},t}=S(t,I_{\mathrm{target}})-S(t-1,I_{\mathrm{target}}),
\end{equation}

where $r_{\mathrm{coarse},0}$ is set to $0$. Once the current observation is sufficiently close to the target frame, the target index is advanced by a fixed stride $\Delta I$:

\begin{equation}
I_{\mathrm{target}} \leftarrow
\left\{
\begin{array}{ll}
\min(I_{\mathrm{target}}+\Delta I,l), & S(t,I_{\mathrm{target}})>\tau_c,\\
I_{\mathrm{target}}, & \mathrm{otherwise}.
\end{array}
\right.
\end{equation}

The second component is a fine-grained matching reward, which rewards accurate visual alignment with any sufficiently advanced guidance frame. For each observation, we first find the most similar generated frame:

\begin{equation}
i_f(t)=\arg\max_{i\in\{1,\ldots,l\}}S(t,i).
\end{equation}

To avoid repeatedly rewarding the same early guidance state, DEG maintains the rearmost rewarded index $I_{\mathrm{reached}}$, initialized as $0$. A fine-grained reward is assigned only when the matched frame is confident enough and goes beyond the previously rewarded progress:

\begin{equation}
r_{\mathrm{fine},t} =
\left\{
\begin{array}{ll}
i_f(t), & S(t,i_f(t))>\tau_f \ \mathrm{and}\ i_f(t)>I_{\mathrm{reached}},\\
0, & \mathrm{otherwise}.
\end{array}
\right.
\end{equation}

The progress record is updated as

\begin{equation}
I_{\mathrm{reached}} \leftarrow
\left\{
\begin{array}{ll}
\max(I_{\mathrm{reached}},i_f(t)), & r_{\mathrm{fine},t}>0,\\
I_{\mathrm{reached}}, & \mathrm{otherwise}.
\end{array}
\right.
\end{equation}

Finally, the DEG-style exploration incentive used by VLMeg combines the coarse and fine rewards with coefficients $\alpha$ and $\beta$:

\begin{equation}
r_{\mathrm{expl},t}=\alpha r_{\mathrm{coarse},t}+\beta r_{\mathrm{fine},t}.
\end{equation}

\clearpage
\subsection{B.2. VLMeg Pseudo Code}

In this section, we provide the pseudo code of VLMeg in Algorithm~\ref{alg:vlmeg}. The algorithm summarizes the full reward-construction pipeline, including episode-conditioned success-reference generation, online trajectory filtering, key-observation selection, VLM success verification, and reward combination for policy learning.

\begin{algorithm}[h]
\caption{The pseudo code of the proposed VLMeg.}
\label{alg:vlmeg}
\begin{algorithmic}[1]
\REQUIRE Task prompt $q$, expert videos $\mathcal{V}$, image-to-video generator $G$, frozen VLM detector $D$, RL algorithm $\mathcal{A}$, thresholds $\tau_{\mathrm{idx}}$ and $\tau_{\mathrm{sim}}$, success-reward weight $\lambda$
\ENSURE Learned policy $\pi_\theta$
\STATE Adapt the generator with LoRA: $G_\psi \leftarrow \mathrm{LoRA}(G;\mathcal{V},q)$
\STATE Train the task-specific similarity encoder $E_\phi$ on $\mathcal{V}$ with the contrastive self-supervised objective
\STATE Initialize policy $\pi_\theta$ and replay buffer $\mathcal{B}$
\FOR{each RL episode $e$}
    \STATE Observe the initial frame $o_0^e$
    \STATE Generate episode-conditioned guidance:
    $\tilde{V}^e=(\tilde{v}_1^e,\ldots,\tilde{v}_l^e) \leftarrow G_\psi(o_0^e,q)$
    \STATE Build the success-reference frame set:
    $\mathcal{I}_+^e \leftarrow \{i\in\{1,\ldots,l\}\mid i\geq\tau_{\mathrm{idx}}\}$,
    $\mathcal{F}_+^e \leftarrow \{\tilde{v}_i^e\mid i\in\mathcal{I}_+^e\}$
    \STATE Set $\mathrm{worthy}\leftarrow \mathrm{false}$
    \FOR{$t=0$ to $T$}
        \STATE Compute success-region similarity:
        $s_t \leftarrow \max_{i\in\mathcal{I}_+^e}\cos(E_\phi(o_t^e),E_\phi(\tilde{v}_i^e))$
        \STATE Compute the DEG-style exploration incentive $r_{\mathrm{expl},t}$ using $\tilde{V}^e$ and $E_\phi$
        \IF{$s_t \geq \tau_{\mathrm{sim}}$}
            \STATE $\mathrm{worthy}\leftarrow \mathrm{true}$
        \ENDIF
        \IF{$t<T$}
            \STATE Sample action $a_t^e\sim\pi_\theta(\cdot\mid o_t^e)$ and step the environment to obtain $o_{t+1}^e$
        \ENDIF
    \ENDFOR
    \IF{$\mathrm{worthy}$}
        \STATE Select the key observation: $t^\star \leftarrow \arg\max_{t\in\{0,\ldots,T\}}s_t$
        \STATE Set success reference $v_{\mathrm{ref}}^e\leftarrow \tilde{v}_l^e$
        \STATE Query the VLM once: $b(\tau^e)\leftarrow D(q,o_0^e,o_{t^\star}^e,v_{\mathrm{ref}}^e)$
        \STATE Set episodic success reward $R_{\mathrm{succ}}(\tau^e)\leftarrow b(\tau^e)$
    \ELSE
        \STATE Skip VLM evaluation and set $R_{\mathrm{succ}}(\tau^e)\leftarrow 0$
    \ENDIF
    \FOR{$t=0$ to $T$}
        \STATE Construct VLMeg reward:
        $r_t^{\mathrm{VLMeg}}\leftarrow r_{\mathrm{expl},t}+\lambda R_{\mathrm{succ}}(\tau^e)$
    \ENDFOR
    \STATE Store the relabeled episode with rewards $\{r_t^{\mathrm{VLMeg}}\}_{t=0}^{T}$ in $\mathcal{B}$
    \STATE Update $\pi_\theta$ with RL algorithm $\mathcal{A}$ using minibatches from $\mathcal{B}$
\ENDFOR
\RETURN $\pi_\theta$
\end{algorithmic}
\end{algorithm}

\newpage
\section{C. Detailed Experimental Settings}
\subsection{C.1. Introduction of Environment}

\begingroup
\setlength{\tabcolsep}{10pt}
\begin{center}
\begin{tabular}{@{}m{0.13\columnwidth} m{0.3\columnwidth} m{0.13\columnwidth} m{0.3\columnwidth}@{}}
\includegraphics[width=\linewidth]{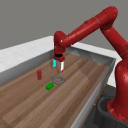}
&
{\raggedright\small
\textit{assembly.}\\
Picking up a nut and placing it onto a peg. The initial positions of both the nut and the peg are randomized, challenging grasping, spatial alignment, and precise placement.
}
&
\includegraphics[width=\linewidth]{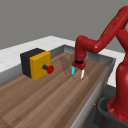}
&
{\raggedright\small
\textit{button-press.}\\
Pressing a button from a randomized initial position. The task tests whether the robot can localize the button and execute a controlled pushing motion.
}
\end{tabular}
\end{center}
\smallskip
\begin{center}
\begin{tabular}{@{}m{0.13\columnwidth} m{0.3\columnwidth} m{0.13\columnwidth} m{0.3\columnwidth}@{}}
\includegraphics[width=\linewidth]{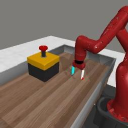}
&
{\raggedright\small
\textit{button-press-topdown.}\\
Pressing a button from the top with randomized button positions. The task requires vertical alignment and accurate downward contact.
}
&
\includegraphics[width=\linewidth]{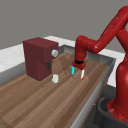}
&
{\raggedright\small
\textit{coffee-button.}\\
Pushing the button on a coffee machine from randomized machine positions. The task requires identifying the button on the appliance and completing a precise press.
}
\end{tabular}
\end{center}
\smallskip
\begin{center}
\begin{tabular}{@{}m{0.13\columnwidth} m{0.3\columnwidth} m{0.13\columnwidth} m{0.3\columnwidth}@{}}
\includegraphics[width=\linewidth]{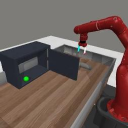}
&
{\raggedright\small
\textit{door-close.}\\
Closing a door mounted on a revolving hinge joint with randomized door positions. The task requires adapting the pushing trajectory to the door geometry.
}
&
\includegraphics[width=\linewidth]{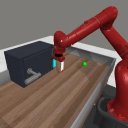}
&
{\raggedright\small
\textit{door-open.}\\
Opening a door with a revolving hinge joint under randomized door positions. The robot must reach the handle and pull the door through a successful opening motion.
}
\end{tabular}
\end{center}
\smallskip
\begin{center}
\begin{tabular}{@{}m{0.13\columnwidth} m{0.3\columnwidth} m{0.13\columnwidth} m{0.3\columnwidth}@{}}
\includegraphics[width=\linewidth]{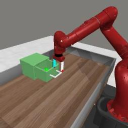}
&
{\raggedright\small
\textit{drawer-close.}\\
Pushing and closing a drawer from randomized drawer positions. The task requires contact with the drawer handle and sustained motion until the drawer reaches the closed state.
}
&
\includegraphics[width=\linewidth]{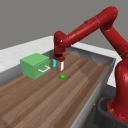}
&
{\raggedright\small
\textit{drawer-open.}\\
Opening a drawer from randomized drawer positions. The robot must grasp or pull the handle and move the drawer outward to the target open state.
}
\end{tabular}
\end{center}
\smallskip
\begin{center}
\begin{tabular}{@{}m{0.13\columnwidth} m{0.3\columnwidth} m{0.13\columnwidth} m{0.3\columnwidth}@{}}
\includegraphics[width=\linewidth]{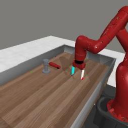}
&
{\raggedright\small
\textit{faucet-close.}\\
Rotating a faucet handle clockwise from randomized faucet positions. The task requires making contact with the handle and completing the closing rotation.
}
&
\includegraphics[width=\linewidth]{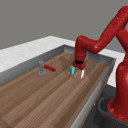}
&
{\raggedright\small
\textit{faucet-open.}\\
Rotating a faucet handle counter-clockwise from randomized faucet positions. The task requires precise contact and sustained rotation to the target open state.
}
\end{tabular}
\end{center}
\smallskip
\begin{center}
\begin{tabular}{@{}m{0.13\columnwidth} m{0.3\columnwidth} m{0.13\columnwidth} m{0.3\columnwidth}@{}}
\includegraphics[width=\linewidth]{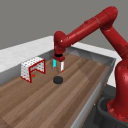}
&
{\raggedright\small
\textit{plate-slide.}\\
Sliding a plate into a cabinet with randomized plate and cabinet positions. The task requires pushing the plate along the correct direction into the target region.
}
&
\includegraphics[width=\linewidth]{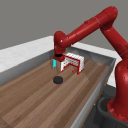}
&
{\raggedright\small
\textit{plate-slide-side.}\\
Sliding a plate into a cabinet sideways with randomized plate and cabinet positions. The task emphasizes lateral pushing and accurate final placement.
}
\end{tabular}
\end{center}
\smallskip
\begin{center}
\begin{tabular}{@{}m{0.13\columnwidth} m{0.3\columnwidth} m{0.13\columnwidth} m{0.3\columnwidth}@{}}
\includegraphics[width=\linewidth]{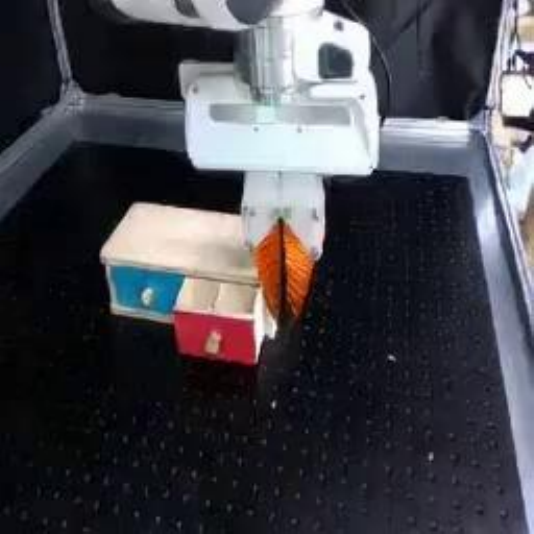}
&
{\raggedright\small
\textit{real-world: drawer-close.}\\
Closing a red drawer in the real-world Franka setup from an open state. The task tests reward learning under real visual observations, occlusion, and contact-rich object motion.
}
&
\includegraphics[width=\linewidth]{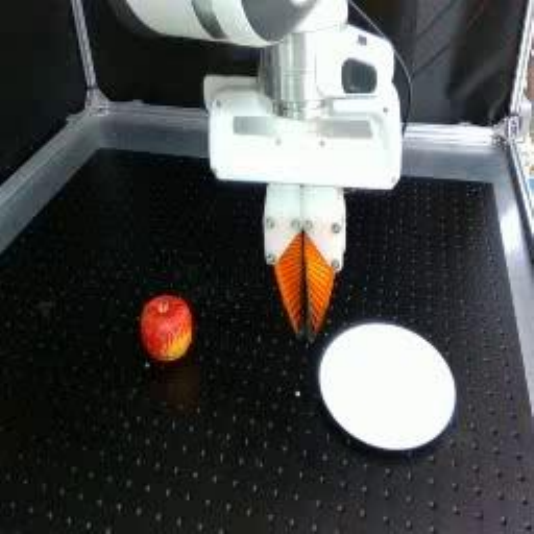}
&
{\raggedright\small
\textit{real-world: pick-apple.}\\
Picking up an apple and placing it into a plate in the real-world Franka setup. The task requires reliable grasping, transport, and placement under visual variation.
}
\end{tabular}
\end{center}
\smallskip
\begin{center}
\begin{tabular}{@{}m{0.13\columnwidth} m{0.3\columnwidth} m{0.13\columnwidth} m{0.3\columnwidth}@{}}
\includegraphics[width=\linewidth]{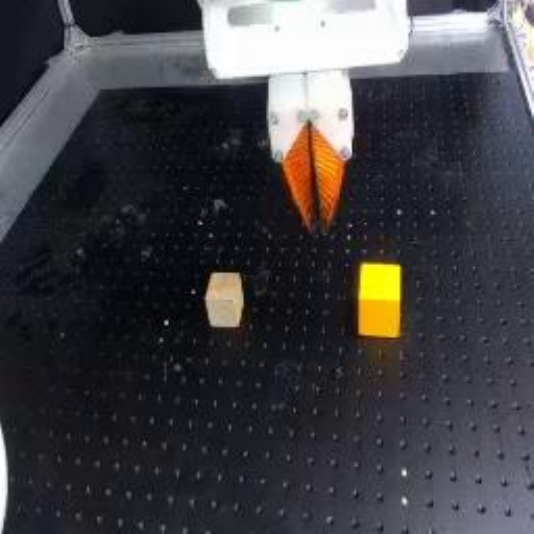}
&
{\raggedright\small
\textit{real-world: stack-cube.}\\
Stacking a yellow cube on top of a white cube in the real-world Franka setup. The task requires accurate object-object contact and stable final placement.
}
&
\mbox{}
&
\mbox{}
\end{tabular}
\end{center}
\endgroup

\subsection{C.2. Baseline Introduction}
In simulation tasks, we compare VLMeg with five representative VLM-based reward baselines: Robo-Reward \cite{roboreward}, Robometer \cite{robometer}, VLAC \cite{vlac}, Robo-Dopamine \cite{robodopamine}, and Dopamine-FT.

\subsubsection{Robo-Reward} \cite{roboreward} is a trajectory-level reward model. It constructs a labeled dataset and trains a VLM-based reward model to directly predict a discrete progress score from 1 to 5. During evaluation, it takes a downsampled episode as input and outputs a task-progress reward for the whole trajectory.

\subsubsection{Robometer} \cite{robometer} follows the trajectory-level reward formulation and further introduces pairwise trajectory comparison during training. This relative preference supervision improves progress estimation by learning from comparisons between trajectory clips. Robometer is handled separately in our experiments because its official RL pipeline samples previous trajectory clips and computes rewards during training. For efficiency, we set its reward interval to 2, meaning that Robometer reward is evaluated every two environment steps.

\subsubsection{VLAC} \cite{vlac} is a frame-pair reward model. It fine-tunes a model to judge the relative progress between two frames, and uses the accumulated progress over adjacent frame pairs as the reward of the current state.

\subsubsection{Robo-Dopamine} \cite{robodopamine} is also a frame-pair reward model. It improves frame-pair progress estimation with three complementary inference modes based on the initial frame, the goal frame, and adjacent frames. To fully evaluate its reward quality, we average the rewards predicted by the three modes as the final state reward.

\subsubsection{Dopamine-FT} \cite{robodopamine} is a task-specific fine-tuned variant of Robo-Dopamine \cite{robodopamine}. We construct a manually annotated dataset from a few successful trajectories and fine-tune Robo-Dopamine on this dataset, yielding Dopamine-FT.

For Robo-Reward, VLAC, Robo-Dopamine, and Dopamine-FT, we use the learned rewards during RL as follows. Once an episode is marked as \texttt{done}, we query each reward model to compute the reward of the current episode or state, and then assign this scalar reward to every transition in the episode.

\clearpage
\subsection{C.3. Hyper-parameter Settings}

We provide the hyper-parameters of VLMeg, as shown in \ref{hyper}. The project code will be publicly released upon the acceptance of our paper.

\begin{table}[h]
\renewcommand\arraystretch{1.15}  
\caption{ The hyper-parameters of VLMeg. }
\centering
\setlength{\tabcolsep}{3mm}{
\begin{tabular}{|lc|}
\hline
Hyper-parameter                      & Setting       \\ \hline
Video fondation Model & Wan2.1-I2V-14B \\
Number of finetuning videos &  3-5 \\
Video frames & $250$ \\
Frame shape & $256\times256\times3$ \\
Frame stack & $1$\\
Action repeat & $2$\\
Action type & Continuous \\
Action dimension & See Metaworld paper\\
Large model finetuning & LoRA\\

self-supervised pre-training times & $5000$ \\
Feature encoder lr & $1e-4$ \\
Random shift (noise) upper bound& $4$ \\
EMA update frequency & $2$ \\
EMA Momentum & 0.05 \\
Optimizer & Adam \cite{adam} \\
VLM backbone  &  GPT-5.5  \\
Hyper-parameter settings of $r_{\mathrm{expl},t}$ & Default in DEG paper\\

Success reward weight $\lambda$ & 10 \\
Similarity threshold $\tau_{\mathrm{sim}}$ & 0.95 \\

Metaworld RL backbone & DrQv2 \\
Discount factor   & $0.98$          \\
Batch size & $256$ \\
n-step returns  & $3$ \\
Random frames in RL & $4000$ \\
Feature dimension  & $50$ \\
Task difficulty for exploration & Hard \\

Other DrQv2 hyper-parameters & See DrQv2 paper \\

Real-world RL backbone & HIL-SERL \\
HIL-SERL hyper-parameters & See HIL-SERL paper \\

\hline
\end{tabular}
}
\label{hyper}
\end{table}

\clearpage
\subsection{C.4. VLM Prompt Details in VLMeg}

\subsubsection{Prompt framework.}
All task prompts follow a shared three-image comparison structure. The prompt first describes the task-relevant objects and the desired object-state change. It then asks the VLM to compare the first two reference images, where the first image shows the state before task completion and the second image shows the state after task completion. Finally, the VLM is asked to judge whether the third image corresponds to the completed state. Across tasks, the prompt explicitly instructs the VLM to ignore absolute target positions and focus on task-relevant state changes; for tasks where the robot arm may distract the judgment, the prompt also asks the VLM to ignore the robot-arm position. The output format is strictly binary: output $1$ only for task completion and output $0$ otherwise, without any additional text.

\newcommand{\vlmprompt}[2]{%
\par\medskip
\noindent\rule{\linewidth}{0.35pt}\par
\noindent\textbf{#1}\par
\noindent\textit{Prompt.} #2\par
}

\vlmprompt{\texttt{assembly}.}{The image contains a gray ring attached to a green block and a separate red cylindrical workpiece. The task is to assemble the ring onto the workpiece. Please analyze the difference between the first two images: the first image corresponds to the state before task completion, and the second image corresponds to the state after task completion. Please determine whether the third image is before or after task completion. Ignore the absolute target position and focus only on the target state change. If the task is completed, output $1$; otherwise, output $0$. Do not output anything else.}

\vlmprompt{\texttt{button-press}/\texttt{button-press-topdown}.}{The image contains a yellow block with a red button on it. The task is to fully press the red button. After task completion, the red button and the yellow block change from being connected through a black post to being in direct contact. Please analyze the difference between the first two images: the first image corresponds to the state before task completion, and the second image corresponds to the state after task completion. The black post may become shorter while the red button is still not in direct contact with the yellow block; this case should still be treated as before task completion. Please determine whether the third image is before or after task completion. Ignore the absolute target position, focus only on the target state change, and ignore the robot-arm position. If the task is completed, output $1$; otherwise, output $0$. Do not output anything else.}

\vlmprompt{\texttt{coffee-button}.}{The image contains a reddish-brown coffee machine with a white button on it. The task is to press the white button. Please analyze the difference between the first two images: the first image corresponds to the state before task completion, and the second image corresponds to the state after task completion. Please determine whether the third image is before or after task completion. Ignore the absolute target position, focus only on the target state change, and ignore the robot-arm position. If the task is completed, output $1$; otherwise, output $0$. Do not output anything else.}

\vlmprompt{\texttt{door-close}.}{The image contains a black door. The task is to fully close the door. Please analyze the difference between the first two images: the first image corresponds to the state before task completion, and the second image corresponds to the state after task completion. Please determine whether the third image is before or after task completion. Ignore the absolute target position, focus only on the target state change, and ignore the robot-arm position. If the task is completed, output $1$; otherwise, output $0$. Do not output anything else.}

\vlmprompt{\texttt{door-open}.}{The image contains a black door. The task is to fully open the door. Please analyze the difference between the first two images: the first image corresponds to the state before task completion, and the second image corresponds to the state after task completion. If the door is only partially open and has not reached the fully open state shown in the second image, it should still be treated as before task completion. Please determine whether the third image is before or after task completion. Ignore the absolute target position, focus only on the target state change, and ignore the robot-arm position. If the task is completed, output $1$; otherwise, output $0$. Do not output anything else.}

\vlmprompt{\texttt{drawer-close}.}{The image contains a green drawer with a white handle. The task is to close the drawer. Please analyze the difference between the first two images: the first image corresponds to the state before task completion, and the second image corresponds to the state after task completion. Please determine whether the third image is before or after task completion. Ignore the absolute target position, focus only on the target state change, and ignore the robot-arm position. If the task is completed, output $1$; otherwise, output $0$. Do not output anything else.}

\clearpage
\vlmprompt{\texttt{drawer-open}.}{The image contains a green drawer with a white handle. The task is to open the drawer by pulling the white handle. Please analyze the difference between the first two images: the first image corresponds to the state before task completion, and the second image corresponds to the state after task completion. Please determine whether the third image is before or after task completion. Ignore the absolute target position, focus only on the target state change, and ignore the robot-arm position. If the task is completed, output $1$; otherwise, output $0$. Do not output anything else.}

\vlmprompt{\texttt{faucet-close}.}{The image contains a gray cylinder, which is the valve, and a red elongated object attached to it, which is the valve handle. The task is to close the valve by rotating the handle. Please analyze the difference between the first two images: the first image corresponds to the state before task completion, and the second image corresponds to the state after task completion. Please determine whether the third image is before or after task completion. Ignore the absolute target position and focus only on the target state change. If the task is completed, output $1$; otherwise, output $0$. Do not output anything else.}

\vlmprompt{\texttt{faucet-open}.}{The image contains a gray cylinder, which is the valve, and a red elongated object attached to it, which is the valve handle. The task is to open the valve by rotating the handle. Please analyze the difference between the first two images: the first image corresponds to the state before task completion, and the second image corresponds to the state after task completion. Please determine whether the third image is before or after task completion. Ignore the absolute target position and focus only on the target state change. If the task is completed, output $1$; otherwise, output $0$. Do not output anything else.}

\vlmprompt{\texttt{plate-slide}/\texttt{plate-slide-side}.}{The image contains a red-and-white door-shaped frame, and there is a small red dot on the floor inside the frame marking the target position. The task is to move a black circular flat cylindrical object to the red-dot position. Please analyze the difference between the first two images: the first image corresponds to the state before task completion, and the second image corresponds to the state after task completion. Please determine whether the third image is before or after task completion. Ignore the absolute target position, focus only on the target state change, and ignore the robot-arm position. If the task is completed, output $1$; otherwise, output $0$. Do not output anything else.}

\vlmprompt{\texttt{real-world: drawer-close}.}{The image contains a white cabinet with one red drawer and one blue drawer. The task is to close the red drawer. Please analyze the difference between the first two images: the first image corresponds to the state before task completion, and the second image corresponds to the state after task completion. Please determine whether the third image is before or after task completion. Ignore the absolute target position, focus only on the target state change, and ignore the robot-arm position. If the task is completed, output $1$; otherwise, output $0$. Do not output anything else.}

\vlmprompt{\texttt{real-world: pick-apple}.}{The image contains a red apple and a white plate. The task is to move the red apple into the white plate. Please analyze the difference between the first two images: the first image corresponds to the state before task completion, and the second image corresponds to the state after task completion. Please determine whether the third image is before or after task completion. Ignore the absolute target position, focus only on the target state change, and ignore the robot-arm position. If the task is completed, output $1$; otherwise, output $0$. Do not output anything else.}

\vlmprompt{\texttt{real-world: stack-cube}.}{The image contains a large yellow cube and a small white cube. The task is to move the small white cube onto the large yellow cube. Please analyze the difference between the first two images: the first image corresponds to the state before task completion, and the second image corresponds to the state after task completion. Please determine whether the third image is before or after task completion. Ignore the absolute target position, focus only on the target state change, and ignore the robot-arm position. If the task is completed, output $1$; otherwise, output $0$. Do not output anything else.}
\newpage

\subsection{C.5. Dataset for Motivation Study}
\label{sec:motivation-dataset}
\label{sec:reference_examples}

\providecommand{\motivationinclude}[2]{%
\IfFileExists{#2}{\includegraphics[width=#1]{#2}}{\fbox{\parbox{#1}{\centering Missing figure file}}}%
}

\begin{figure}[!htbp]
\centering
\motivationinclude{0.95\linewidth}{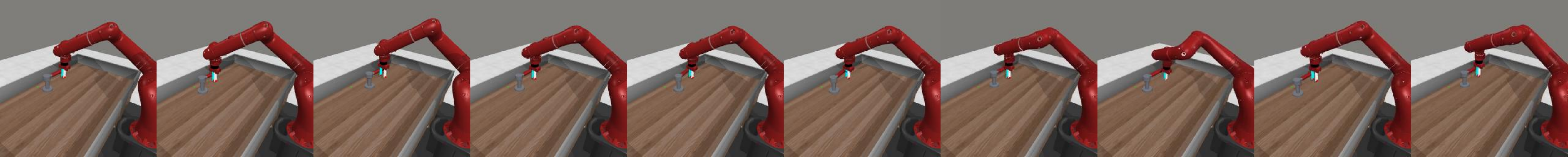}
\par\smallskip
\motivationinclude{0.95\linewidth}{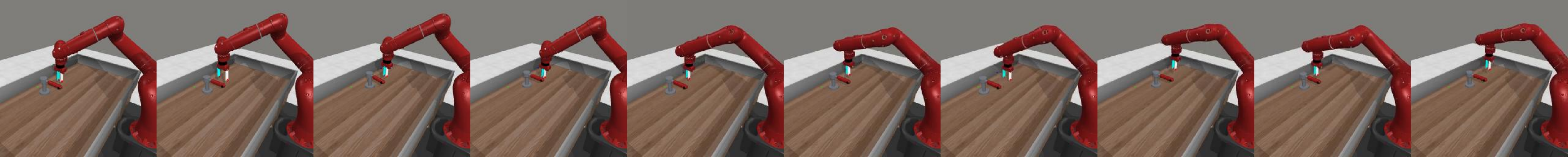}
\caption{Examples in the Near dataset used in the reference-frame motivation study on \texttt{faucet-open}. Top: successful examples close to the expert success frame. Bottom: failing examples whose robot-arm states are highly similar to the expert success frame.}
\label{fig:motivation-near-examples}
\label{fig:motivation-near-success}
\label{fig:motivation-near-fail}
\end{figure}

In Table~\ref{tab:context-motivation}, we compare VLM detection accuracy on \texttt{faucet-open} with and without reference frames. The evaluation dataset contains 20 images. The successful subset contains 10 frames that are visually close to the expert success frame. The failing subset contains 10 frames whose robot-arm states are also highly similar to the expert success frame, yet the target object remains untouched. These successful and failing subsets are shown in Figure~\ref{fig:motivation-near-examples}. In the 'w/o Reference Frames' setting, the VLM makes judgments solely based on task descriptions and target images. In the 'w/ Reference Frames' setting, the VLM input additionally incorporates the initial and success reference frames from expert trajectory, providing richer contextual information and leading to substantial accuracy improvements. 

\begin{figure}[!htbp]
\centering
\motivationinclude{0.95\linewidth}{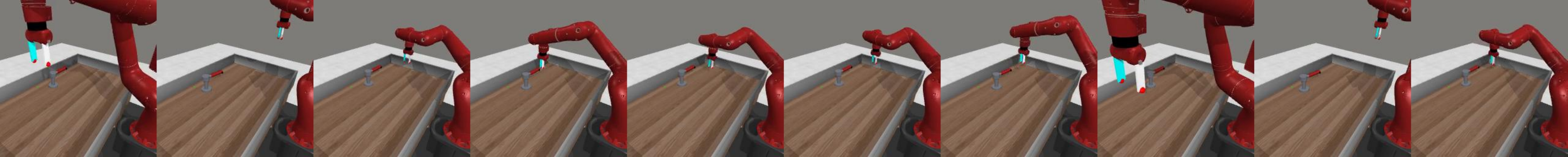}
\par\smallskip
\motivationinclude{0.95\linewidth}{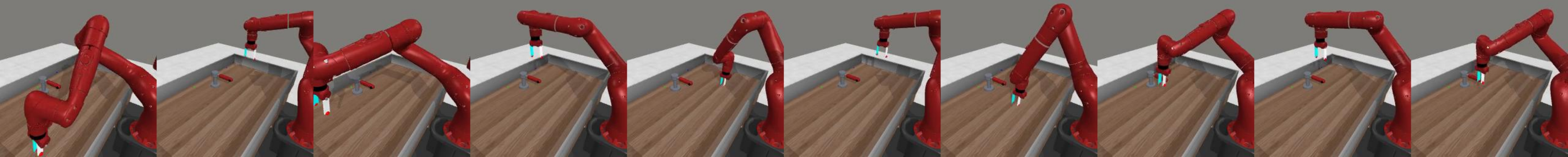}
\caption{Examples in the Deviated dataset. Top: successful examples. Bottom: failing examples. In both subsets, the robot-arm states deviate from the expert success frame, and no occlusion is present.}
\label{fig:motivation-deviated-examples}
\label{fig:motivation-deviated-success}
\label{fig:motivation-deviated-fail}
\end{figure}

In Table~\ref{tab:key-observation-motivation}, the Near dataset is the same 20-image dataset used in Table~\ref{tab:context-motivation}. The Deviated dataset contains 10 successful frames and 10 failing frames whose robot-arm states deviate from the expert success frame. None of these images contain occlusion, so this comparison isolates the influence of robot-arm pose deviation. The successful and failing subsets of the Deviated dataset are shown in Figure~\ref{fig:motivation-deviated-examples}.

\end{document}